# Fault Tolerant Control of a Quadcopter using Reinforcement Learning

Muzaffar Habib [1], Adnan Maqsood[1], Adnan Fayyaz ud Din[2]

[1] National University of Sciences & Technology, Islamabad, Pakistan, 44000

[2] Air University, Islamabad, Pakistan, 44000

*Abstract:* This study presents a novel reinforcement learning (RL)-based control framework aimed at enhancing the safety and robustness of the quadcopter, with a specific focus on resilience to in-flight one propeller failure. Addressing the critical need of a robust control strategy for maintaining a desired altitude for the quadcopter to safe the hardware and the payload in physical applications. The proposed framework investigates two RL methodologies Dynamic Programming (DP) and Deep Deterministic Policy Gradient (DDPG), to overcome the challenges posed by the rotor failure mechanism of the quadcopter. DP, a model-based approach, is leveraged for its convergence guarantees, despite high computational demands, whereas DDPG, a model-free technique, facilitates rapid computation but with constraints on solution duration. The research challenge arises from training RL algorithms on large dimension and action domains. With modifications to the existing DP and DDPG algorithms, the controllers were trained not only to cater for large continuous state and action domain and also achieve a desired state after an inflight propeller failure. To verify the robustness of the proposed control framework, extensive simulations were conducted in a MATLAB environment across various initial conditions and underscoring their viability for mission-critical quadcopter applications. A comparative analysis was performed between both RL algorithms and their potential for applications in faulty aerial systems.

## I. Introduction

Small-scale quadcopters have become widely popular due to their versatility and range of applications in fields like communication, exploration, aerial photography, agriculture, payload transport, and surveillance. This success is largely attributed to advances in lightweight, affordable control systems, GPS, and camera technologies. Extensive research has focused on enhancing the multi-role capabilities of quadcopters through improved modeling and control strategies. Traditionally, many control approaches rely on linearizing the system dynamics around an equilibrium point. While effective in some applications, linearization limits adaptability, especially for systems operating in complex, dynamic environments. Nonlinear control strategies, such as feedback linearization [1][2], sliding mode control [3], backstepping [4], and dynamic inversion [5], offer improved performance but require a comprehensive understanding of system dynamics and state-space models. These controllers often necessitate significant human intervention to adapt to changes in the environment or quadcopter parameters, which can reduce their effectiveness in unexpected failure scenarios. To address the limitations of traditional control strategies, particularly in fault scenarios, adaptive and intelligent control methods are essential. Reinforcement Learning (RL) offers a promising solution, providing self-learning controllers that adaptively respond to environmental changes and system dynamics [6]. Model-free RL algorithms are especially advantageous in scenarios where precise system models are unavailable, allowing the RL agent to learn the dynamics through interactions with the environment. These interactions yield scalar rewards that measure the agent's deviation from desired states, guiding it to optimize control actions based on a predefined reward function.

This research addresses this gap by proposing an RL-based control strategy specifically designed to stabilize and control a quadcopter experiencing in-flight propeller failure. Quadcopters have 12 continuous states and 4-dimensional action space. So far RL algorithm has demonstrated large success in discrete state and action domains, however, to expand the application of RL algorithms on physical systems requires more research work. This paper aims at addressing the challenges arising from training actor and critic neural networks on large dimensional systems.

The application of RL to quadcopter control has gained attention in recent years, particularly for managing non-linear dynamics in unmanned aerial vehicles. RL-based controllers have been implemented for hovering, waypoint navigation, and other autonomous flight tasks. For example, Morrison and Fisher [7] developed a neural network-based controller for quadcopter hovering and landing, while studies utilizing Generalized Advantage Estimation (GAE) [8] and Proximal Policy Optimization (PPO) [9] have shown promising results for UAV control. Other applications include training quadcopters for complex tasks via neural networks [10], developing RL-based autopilots for attitude control [11], and using RL for executing complex maneuvers in helicopters [12].

Quadcopters are particularly vulnerable to single propeller failure, which results in an asymmetric power configuration. Such a failure immediately generates rolling and pitching moments, causing the quadcopter to topple. The situation is further complicated by the yawing moment created by the torque differential of the remaining operational rotors. Unlike traditional controllers that require predefined control structures and may struggle with compromised control authority, the RL controller adapts dynamically, leveraging state-action experiences to maintain hover and altitude control with only three functional propellers. Table 1 summarizes various techniques applied to fault-tolerant control in quadcopters. While much research has focused on designing RL controllers for autonomous flight, limited work addresses post-failure stabilization and control, especially in scenarios where control authority is compromised.

.

Table 1 : Fault detection and control strategies of Quadcopter

| S No | Year | Author | Fault | Conference/ Journal | Methodology | Remarks | Ref |
|---|---|---|---|---|---|---|---|
| 1. | 2020 | Gerardo Ortiz-Torres et al | Actuator Fault | IEEE Access | Quasi Linear Parameter Varying | 60 secs non-hover Flight | [19] |
| 2. | 2020 | Ngoc Phi Nguyen, Nguyen Xuan Mung | Reduced thrust from multiple rotor | Mathematics | Neural Network Approximation | Stabilization till 100 secs | [20] |
| 3. | 2019 | N. P. Nguyen and S. K. Hong | Reduced thrust from one rotor | Energies (MDPI) | Normal Adaptive Sliding Mode Control | Trajectory Tracking | [21] |
| 4. | 2014 | W. Mueller and R. D'Andrea | Multiple propeller failure | IEEE Conference on Robotics and Automation | LTI System | Circular trajectory for 20 secs | [22] |
| 5. | 2013 | A. Akhtar, et al | Single Propeller failure | IEEE Conference on Decision and Control | Coordinate and feedback transformation | Constant circular path | [23] |
| 6. | 2011 | A. Freddi, et al | Single Propeller failure | IFAC World Congress Conference | Double Control Loop | Constant yawing flight for 300 secs | [24] |
| 7. | 2008 | Mian, A.A. and D.B Wang | Partial effectiveness of propeller | Journal of Zhejiang University: Science | Exact Feedback Linearization | Quasi stationary flight | [25] |

This research aimed to design a fault-tolerant control strategy for a quadcopter experiencing an in-flight propeller failure. The RL based controller would aim at maximizing the overall hovering time of the damaged propeller so that the hardware can be rescued, and payload can be saved in the event of single rotor failure. The developed controller maintains a specific altitude above the ground using only three propellers instead of the usual four. By leveraging a Reinforcement Learning (RL) algorithm, the controller achieves a hovering state with reduced degrees of freedom, without imposing constraints on orientation angles or x and y location coordinates. The RL controller individually controls the RPMs of the three rotor motors, seeking an optimal action sequence to sustain the quadcopter's hover for the maximum possible duration.

## II.    Modeling & problem formulation

### A.    Quadcopter dynamic model

The quadcopter utilized in this research features four equally spaced rotors, each with independent speed controllers. The representative inertial and body axis system is depicted in Figure 1. Rotors 1 and 2 rotate in the clockwise direction, while rotors 3 and 4 rotate counterclockwise. In a hovering state, all four rotors operate at the same RPM, and the torque produced by opposing pairs of rotors along the z-axis cancels out. A positive rolling motion occurs when rotors 3 and 4 generate the same thrust, and rotor 2 produces slightly more thrust than rotor 1. Positive pitching motion is induced when rotors 1 and 2 generate the same thrust, and rotor 3 produces slightly more thrust than rotor 4. Yawing is controlled by creating a torque differential between the rotors.

To stabilize and control the quadcopter after the loss of one propeller, the RPMs of the remaining three propellers must be carefully managed. This control aims to minimize moments along all three axes while maintaining the quadcopter's altitude.

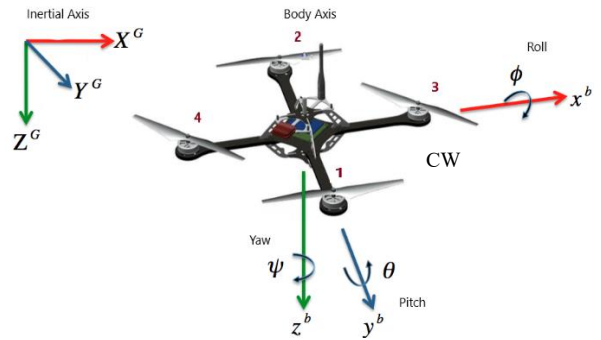

**Figure 1. Quadcopter and axes notation. This figure shows the positive direction of quadcopter axes and rotation direction of rotors**

The input, output and state parameters of any physical model can be modeled as a set of first order differential equations. [26]. Mathematically, these equations are represented as Equation 1.

$$\dot{\mathbf{x}} = \mathbf{f}(\bar{\mathbf{x}}, \bar{\mathbf{u}}) \quad (1)$$

where 'x' is the state vector and 'u' is the control vector. A minimum of 12 states are required to completely represent the rigid body dynamics of any aerial vehicle [27]. For simplicity, a flat non-rotating earth model was used. The state vector used for the modelling of the quadcopter is represented as (Equation 2)

$$\bar{\mathbf{x}} = [U\ V\ W\ P\ Q\ R\ \phi\ \theta\ \psi\ X_n\ Y_n\ Z_n]^T \quad (2)$$

where U, V and W are body axis translational velocities. P, Q and R are body axis angular velocities. Phi ($\phi$), Theta ($\theta$) and Psi ($\psi$) are orientation angles and Xn, Yn and Zn are location coordinates. The input control vector is represented **as** (Equation 3)

$$\bar{\mathbf{u}} = [\Omega_1\ \Omega_2\ \Omega_3\ \Omega_4] \quad (3)$$

$\Omega_1\ \Omega_2\ \Omega_3\ \Omega_4$ are corresponding rotor RPMs.

### B. Calculation of Thrusts and Moments

The thrust mathematical model is governed by an input vector of individual RPMs of each rotor. RL algorithm uses RPMs of three rotors to maintain the altitude. Quadcopter propeller thrust can be calculated directly from rotor RPMs, but for a more accurate thrust model, quadcopter body axis velocities must be used for calculating rotor force. The propeller thrust equation (4), was used for the calculation of thrusts of individual rotors [28], which catered for the

The total z-force and the yawing moment equations remained the same. The rolling moment and pitching moment equation were modified as per the geometry. The new pair of equations gave better control authority to the quadcopter. In the original coordinates, there was no possible condition to produce negative pitching moment, as F4 was zero. The modified set of equations (Table 2) showed that now a condition existed where negative pitching moment could be generated with difference in F2 and sum of F1 and F3. The distances of the rotors were also changed as per modified coordinate axes.

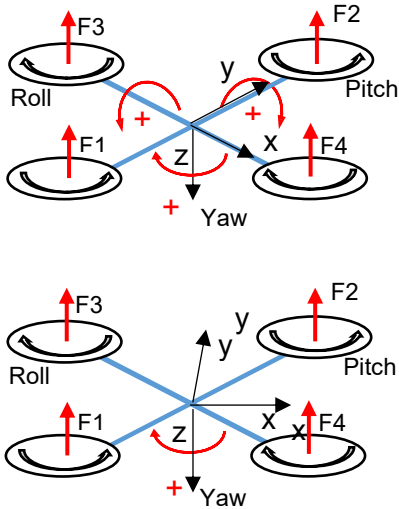

correction in flow properties over the propeller due to quadcopter translation and rotational motions.

$$F_i = \frac{b}{2\pi} \int_0^{2\pi} \int_0^{\Omega_i} \frac{\Delta L\ (\psi, r)}{\Delta R} dr\ d\psi \quad (4)$$

The closed form solution of the integral yields.

$$F_i = \frac{\rho a b c R}{4} \left[ W\ \Omega_i R + \frac{2}{3}(\Omega_i R)^2(\theta_0 + \frac{3}{4}\theta_1) + (U + V)^2(\theta_0 + \frac{1}{2}\theta_1) \right] \quad (5)$$

The final equation (5) used for the calculation of thrusts caters for the correction in flow properties over the propeller due to quadcopter translation and rotational motion. $\Omega$ corresponds to rotor RPM, U, V and W, are quadcopter velocities in body frame. $\theta_0$ & $\theta_1$ are propeller geometric constants. The quadcopter is controlled in all six degrees of freedom via simultaneous control of all four motors. Total thrust of the quadcopter was calculated by summation of individual rotor thrust forces (Equation 6)

$$T = F1 + F2 + F3 + F4 \quad (6)$$

The translational and rotational motion along x and y axes is generated by creating a thrust differential between opposite rotors, whereas altitude is changed by varying thrust of all four rotors collectively. It may be noted that x and y-axis of the quadcopter are tilted in the same plane by 45 degrees as shown in Figure 2. This transformation is done to get control of motors in all three rotational axes in case of single propeller failure. This allows greater control authority along x-y axes using three propellers.

**Figure 2. The figure shows the comparison between standard quadcopter axes and underactuated quadcopter transformed axes**

**Table 2: Quadcopter Force and Moments Equation. This table shows the difference in moment equations using standard axes and with the transformed axes.**

|  | Standard Axes | Transformed Axes |
|---|---|---|
| Fz | - (F1+F2+F3+F4) | - (F1+F2+F3+F4) |
| Rolling Moment (L) | (F2 - F1)*$L_d$ | (F2+F3)*Ly – (F1+F4)*Ly |
| Pitching Moment (M) | (F3 - F4)*$L_d$ | (F1+F3)*Lx – (F2+F4)*Lx |
| Yawing Moment (N) | (F1+F2+F3+F4)*c | (F1 +F2+F3+F4)*c |

**where $L_d$, Lx and Ly are the quadcopter moment arms.**

### III. Research Methodology

A brief overview of the research is provided in te subsequent sections.

## A. Reinforcement learning

Reinforcement learning (RL) has emerged as the most successful and promising Machine Learning paradigm. Reinforcement Learning Algorithms are aimed at optimizing the reward value by carefully adjusting controller behavior via continuous interaction with the environment. RL has found many applications in wide ranging fields such as internet search engines, disease diagnosis, game theory and control.

The value function, denoted as ($V_\pi^*$) represents the expected cumulative reward that an agent can achieve starting from a given state $s$ and following the policy $\pi$. This function is formally defined as

$$V_\pi^*(s = \mathbb{E}_\pi \left[ \sum_{t=0}^{\infty} \gamma^t r_{t+1} \mid s_0 = s \right] \quad (7)$$

Where $r_{t+1}$ denotes the reward at time $t+1$ and $\gamma$ [0, 1] is the discount factor that controls the weighting of future rewards relative to immediate rewards. The action-value function ($Q_\pi^*$) quantifies the expected cumulative reward starting from a specific state $s$, taking an initial action $a$, and subsequently following the policy $\pi$. This function is expressed as

$$Q_\pi^*(s,a) = \mathbb{E}_\pi \left[ \sum_{t=0}^{\infty} \gamma^t r_{t+1} \mid s_0 = s, a_0 = a \right] \quad (8)$$

where $s$ and $a$ represent the initial state and action, respectively. The action-value function thus provides a measure of the "quality" or effectiveness of executing action $a$ in state $s$, with the objective of maximizing long-term rewards by adhering to policy $\pi$ thereafter. It is always a great challenge to choose an appropriate RL algorithm for a specific problem. RL algorithms have been successfully applied to discrete state and action space problems. The implementation of RL algorithm becomes challenging when the state variable values are continuously distributed over large ranges. In this research, two RL algorithms were selected for application on the quadcopter control, Dynamic Programming (MDP) and Deep Deterministic Policy Gradient (DDPG).

## B. RL Definitions

Few basic RL definitions are briefly explained below

**a.    State (s):** State is a physical description of an observation or a measurable quantity. For e.g., position, speed are some states of a moving car.

**b.    RL Agent:**    A RL agent can be understood as the brain of RL algorithm; it takes an action on the plant (quadcopter) and receives updated state vector and reward value. In this problem, the flight controller may be called RL agent.

**c.    Action:**  Actions are a set of all possible actions that can be taken by an agent. In quadcopter control, the input control vector 'u', containing 4 rotor RPMs, can be understood as possible actions for the agent.

**d.    Environment / Plant:**    A plant or an environment is the physical system on which actions are being performed by the agent. In our case, the quadcopter is the plant. It takes actions ($a_t$) from the agent, being in state ($s_t$) and produces state ($s_{t+1}$) and reward as output. For this research, the state includes the position coordinates, linear velocity and angular velocities of the quadcopter.

**e.    Policy**:   A policy is agent's behavior of taking an action (a), being in state(s). A policy can be defined as a set of action taken from initial to final state. For this research, the time history of rotors RPMs is assumed as the policy and the goal is to find an optimal policy that renders quadcopter safe considering inflight fault.

**f.    Reward (R)**    : Reward is a numerical scalar value received by a RL agent for being in state ($s_t$) and taking an action ($a_t$). The reward function explicitly define the goal of the agent. The reward value formulates the good or bad action or experiences in the given environment.

**g.    Model:** A model is the prediction of environment's behaviour. It constitutes of systems transition probabilities ($P_{ss}$). The agent learns the dynamics from the simulated data of episodes and adjust its $\mathcal{P}_{ss}^a$ matrix to achieve its desired objective.

$$\mathcal{P}_{ss}^a = \mathbb{P}\left[ S_{t+1} = s' \mid S_t = s, A_t = a \right] \quad (9)$$

**h.    Value function ($V_s$):**    A value function is a distribution of scalar values over complete state space. It defines how much reward can be extracted from the given state. It accumulates the discounted reward over future time steps t+1, t+2 ….

$$V_\pi(s) = E[R_{t+1} + \gamma R_{t+2} + \gamma^2 R_{t+3} + \cdots \mid S_t = s] \quad (10)$$

Dynamic Programming (DP) and Deep Deterministic Policy Gradient (DDPG) were chosen for their complementary strengths. DP, a model-based approach, guarantees convergence to an optimal solution but is computationally intensive, making it less practical for real-time control. DDPG, a model-free algorithm, is faster and well-suited for continuous state-action spaces, though it relies on neural approximations. DP established a reliable reward structure, while DDPG enabled efficient real-time control, providing both theoretical robustness and practical applicability for quadcopter stabilization.

## C. Dynamic Programming (DP)

Dynamic Programming (DP) method works on the of Bellman's principle of optimality. The main idea behind the model free dynamic programming algorithm is to break down continuous control problems into smaller sub-problems and optimize each time step using optimization using plain recursion [22]. The sole purpose for application of DP code is to find an optimal reward function that allows the hovering state of quadcopter with three rotors. Dynamic Programming is computationally expensive for RL problems, but it is guaranteed to converge and finds an optimal value function when applied to continuous state or action spaces. The dynamic programing code was implemented using equation (7), where at each step all actions were evaluated using a given reward function. The action vector resulting in maximum reward values of the states was then stored to make a complete episodic action sequence, as depicted in Figure (3)

$$V(s = S_t) \leftarrow \max[R_t + \gamma V(S_{t+1})] \quad (11)$$

## D. Reward Function

All reinforcement learning algorithms are aimed to optimize the reward function provided to the agents. In most of the practical problems a scalar reward is calculated using the state and action data at a given time step. Calculation of reward also adds to the computational expense of the RL algorithm, because a reward will be calculated at every time step of the episode. The same reward value helps to construct the action value matrix by assigning weightages to actions earning more reward. The response of the RL agents is carefully handled over the course of different episodes by increasing the probability of positive rewarding action. So the algorithm converges to an optimal action function that results in achievement of required task. In the quadcopter control problem the reward value was calculated using the difference between the desired target state and the attained state after the time step (Equation 8). In location coordinates the target value was chosen to be 'z' value as the same height where propeller failure occurred and target values for all angular rates P, Q and R were set to zero. At first, a reward function was assumed based on required task, but later states and weightages were changed to extend the hovering time of the quadcopter.

$$Reward_t = f\left[z_{target} - z_s, P_{target} - P_s, Q_{target} - Q_s, R_{target} - R\right] \quad (12)$$

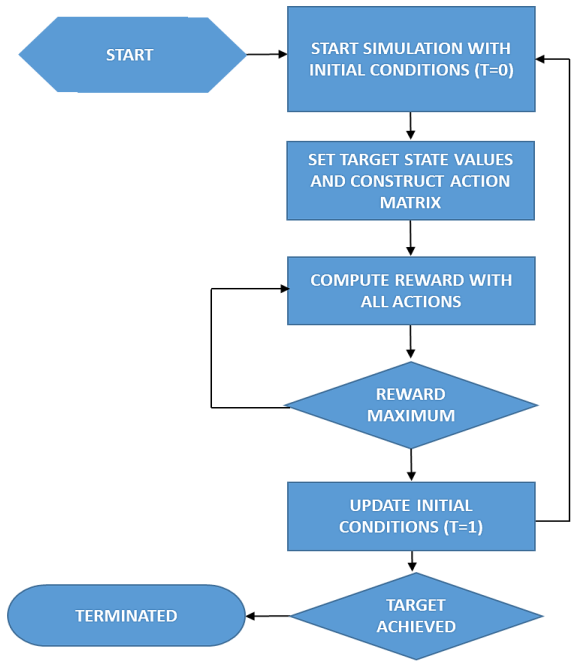

**Figure 3. This figure shows the working principle of Dynamic Programming RL Algorithm.**

### E. *Deep Deterministic Policy Gradient (DDPG)*

The need of constructing and updating a value function over the complete state and action space, becomes a big hurdle in the learning performance of DP RL controller. The solution to this problem is to extract features from the states and use appropriate function approximation technique on those features. The algorithm that uses neural networks as function approximator to update its value function and policy, is widely known as Deep Deterministic Policy Gradient (DDPG). DDPG learns an optimal Action-Value function (Q) and a policy. DDPG is an extension of the DPG algorithm which was introduced by David Silver [29]. Deep neural nets were incorporated by Timothy P. Lillicrap to evaluate deterministic policy gradient [30]. DDPG learns a policy directly from state transition pair and does not need to construct a Q table.

DDPG relies upon optimizing parametrized policies with respect to the expected return by gradient descent. A Stochastic behavior policy is used for good exploration and deterministic target policy for fast convergence. DDPG is an actor-critic algorithm as well; it primarily uses two neural networks, one for the actor, and one for the critic. The actor predicts the action based on the state and the critic predicts the corresponding Q value from a given state action pair. Each neural network returns a probability distribution of selecting an action being in state 's'. In short, the actor takes an action and critic tells it how good or bad the action was. The probability of taking a good action is increased gradually over the course of episodes and RL controller start to exploit the accumulated data of previous episodes (Figure 4). Both the actor and critic neural network parameters are updated simultaneously based on the episodic experiences.

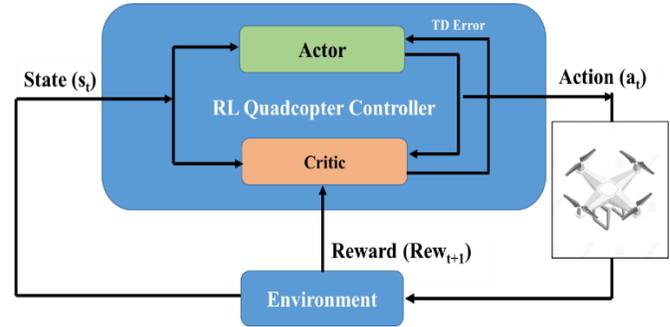

**Figure 4. A working schematic of the DDPG Control Algorithm is presented in this figure.**

DDPG is a well-known RL algorithm famous for application in robotics, game theory etc. However, to make it suitable for flight control applications, a number of techniques were applied to the DDPG algorithm. For example, control of exploration vs. exploitation, actor-critic network, different activation functions, soft target updates, batch normalization, and use of replay buffer. The mentioned techniques are briefly described in subsequent paragraphs.

### F. *Exploration and Exploitation*

One of the major problems in the performance of an RL controller is the tradeoff between exploration and exploitation. This implies that the controller will start to take random actions to get different reward values over complete state space, however over the course of number of episodes, the behavior of the controller will be changed to start exploiting the collected data to yield better rewarding actions. During initial episodes, acting greedily (exploitation) with respect to an approximated function (e.g., Q-function) and choosing the current best action might prevent the agent from discovering new better states and therefore prevent improvement of the policy. On the contrary, excessive exploration might slow down learning or even result in undesired action policies. A tradeoff is therefore necessary between exploration and exploitation. Usually, noise is added to the actions of the RL controller during training. In the case of discrete actions, the $\epsilon$-greedy policy is a common solution: the agent acts randomly with probably $\epsilon$ and greedily with probability $(1 - \epsilon)$. In the case of continuous actions, Gaussian noise could instead be added or through Ornstein-Uhlenbeck process which aids really aids in exploration. But the graph of an OU process is relatively smooth in time, on the other hand, pure Gaussian noise is not temporally correlated and is therefore less suitable for control tasks.

### G. Actor Critic Network (ACN)

The main problem with the policy gradient theorems is finding a suitable score function to evaluate and update policy (Vijay Konda [31]). The same problem led to the evolution of another hybrid algorithm called Actor Critic Method. The algorithm used two different networks, the actor was a function approximator that predicted the action based on current state, and the critic constructed the action value function (Q) based on this state action pair and returned how good or bad the actor action was. The actor learns the behavior of the plant and predicts suitable actions in specific states, while the critic constructs the Q-value function to maximize the reward and achieve desired task. The input and output parameters of Actor and Critic Neural Nets are depicted in Figure 5.

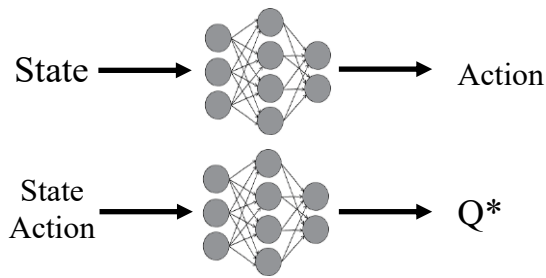

**Figure 5. This figure shows the input and output parameters of Actor Critic Neural Networks**

### H. Activation function

An activation function maps the input data of a neuron to the output. Different activation functions are used with Actor and Critic neural network layers, based on the specific problem. Bin Ding [32] discussed the characteristics of activation functions in deep neural networks. As the actor network predicts the action vector that contains rotor RPMs, a sigmoid was used with final output layer and linear line rectification unit (Relu) for the hidden layers. The same functions were also used for the critic network as well.

### I. Replay Buffers

One paramount challenge in training of neural networks is that most optimization algorithms assume the data samples are Independent and Identically Distributed (IID). This condition is not applicable to our problem, where each state and action is dependent on the state on the previous time step. The quadcopter also generates sequential flight data that limits the learning of the actor-critic neural networks. DDPG uses a novel technique called "Replay Buffer" to mitigate this data problem. Replay buffers store each time step transition data consisting of state ($s_t$), action ($a_t$), next state ($s_{t+1}$), and corresponding scalar reward ($s_t$), which makes 17 values. It stores all transitions in a matrix and uses that cache to update its target parameters [33]. To make optimization more efficient, learning and updating actor critic parameters is done in mini batches. This cache is used to get the effects of complete state and action space. The direction of the stochastic gradient descent is calculated using mini-batch data from replay buffer. DDPG computes gradient after each batch, and updates parameters, instead on taking one-step in descent direction. It takes small steps toward the optimal point with some bias, and this makes the DDPG, computationally stand out from other RL algorithms

### J. Batch normalization

Actor critic neural networks are prone to mean and variance of the sample data, that is quadcopter states history in our problem. The quadcopter was trained to yield different x-location, y-location, z-location, and Euler angles (Phi, theta and psi). The values of these states vary over a wide range as 3 states are locations coordinates in meters and other 3 are Euler angles in radians.

Results from the research of Ioffe & Szegedy [34], proved that batch normalization played an important role in reducing variance and reducing the time and data required for learning. In this research, all 6 states were normalized to values between 0 and 1. This made DDPG much more stable, and it was able to handle large ranges of states and action space data. DDPG resultant values were again multiplied with their weightages to obtain the final state and action values.

### IV. Results and Analysis

The results achieved by Dynamic Programming and Deep Deterministic Policy Gradient algorithm will be discussed in subsequent paragraphs. State history and reward values are plotted against time. Reward values are also plotted with time steps. At first, all states derivatives were set to zero to depict that the quadcopter was in a stable hovering state when the propeller failure occurred. The failure location was set to be 100m above the ground. Later different initial conditions were also tested with the same RL controller.

### A. Dynamic Programming Results

Failure of a propeller in air resulted in an undesired continuous yawing motion. As expected, the quadcopter started rotating in the direction of two running propellers rotation direction. During the formulation of reward function with different weightages of height difference and angular rates, it was observed that quadcopter could not attain target altitude while yaw rate 'R' kept on increasing. The yaw rate problem was carefully handled using reward function switching. Reward function switching is a new term in RL and has not been applied before. Reward function switching is a smart way of changing your desired objective based on some state value. Instead of one reward function, two separate reward functions were used. If the value of yaw rate is within acceptable limits, that's less than 10 rad/sec the controller will try to attain the altitude. As the value of yaw rate exceeds 10 rad/sec the first reward function becomes active, the agent has now liberty to ignore height difference and focus on reducing R value. The controller will momentarily shut down two main lift producing rotors (i.e. 1 and 2) and run the rotor, opposite to failed rotor to maximum RPM to causing a negative moment in 'R' value. During this time the quadcopter loses height as well. After yaw rate values is back within acceptable limits the RL agent will switch reward functions and regains the altitude. The controller does this switching periodically to keep the quadcopter flying at the desired altitude. Mathematically reward function structure is written in equations 9 and 10.

If yaw rate is in acceptable limits:

$$Reward_t = f[z_{target} - z_{t+1}] \quad (13)$$

If yaw rate is above the acceptable limit

$$Reward_t = f[R_{target} - R_{t+1}] \quad (14)$$

The result from the above reward structure with DP code are depicted in Figures 6 to 9. The results showed that the quadcopter was able to hover for 3000 secs i.e. 50 mins. The action sequence generated by the code depicted that the controller gave pulses to the third propeller and operated it at full RPM for 5-to-6-time steps to reduce R and then shut it down to control P and Q. This procedure was iteratively repeated for complete simulation time. The zoomed view of location coordinates and Euler angles history is shown in Figure 9.

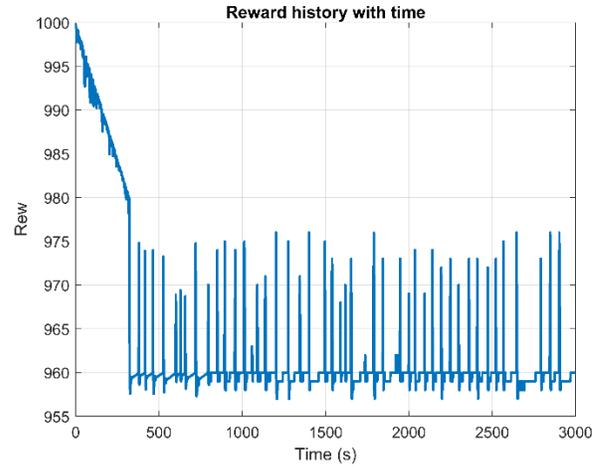

**Figure 6.** This figure shows the variation of reward values for the complete episode. Reward values kept on decreasing till 260 secs due to increase in yawing moment, until reward function switching is activated

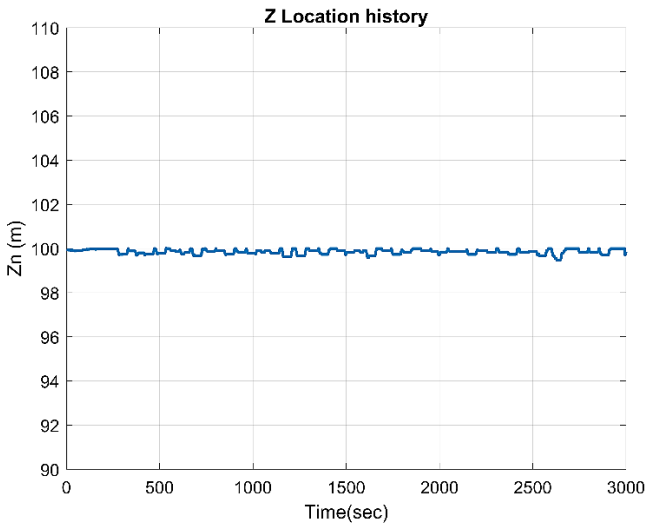

**Figure 7.** This figure shows the variation of z-location coordinate for 3000 secs.

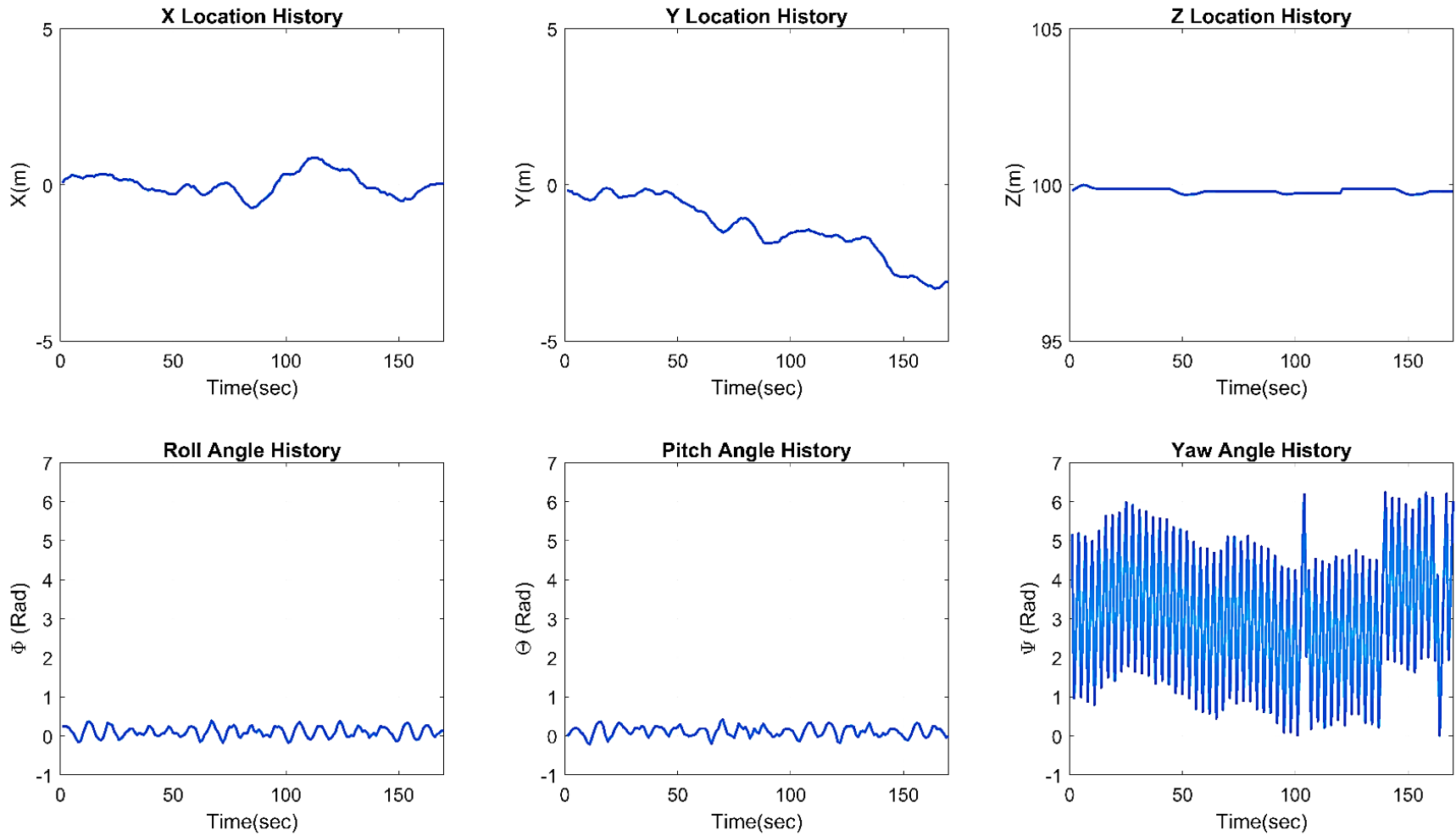

**Figure 8. This figure shows the state's history up to 180 secs using Dynamic Programming Algorithm**

## B. DDPG RL Controller Results

DDPG algorithm works quite differently from the DP algorithm. The controller starts to explore complete action space using random RPM values and results in different quadcopter locations. The length of episode was reduced to 170 secs so that the location of quadcopter can be confined in a small state space. An episode terminates when quadcopter reaches the ground or complete hovering time of 170 secs. The reward value of each step is accumulated to give a final score value to complete the episode. The score value gradually increases during different episodes as the controller gets trained from the experiences. Over 130,000 episodes were required to train the quadcopter for hovering task and coverage a score function to a maximum value. As the memory of replay buffer is filled the actor starts to takes the better rewarding actions and finally reward function converges to a higher range in reward graph (Figure 10). The state history of the best rewarding episode among all 13,0000 episodes is shown in Figure 11.

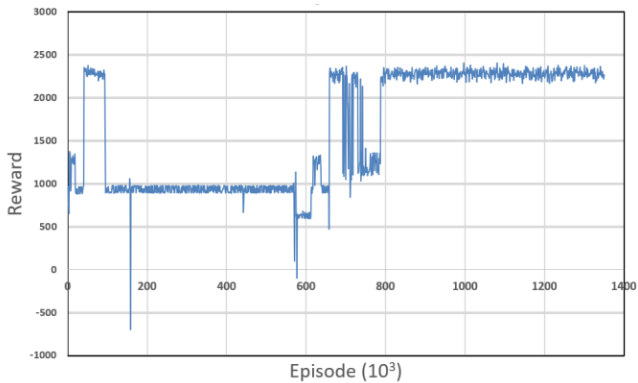

**Figure 9. This figure shows the history of episodic reward gained by DDPG Algorithm during course of learning. The graph depicts the stabilization of reward values between 2200 to 2400 maximum values as the replay buffer updates the neural nets parameters**

The results of quadcopter states are shown in Figure 11. A comparison between DP and DDPG algorithm states is also shown in Figure 12.

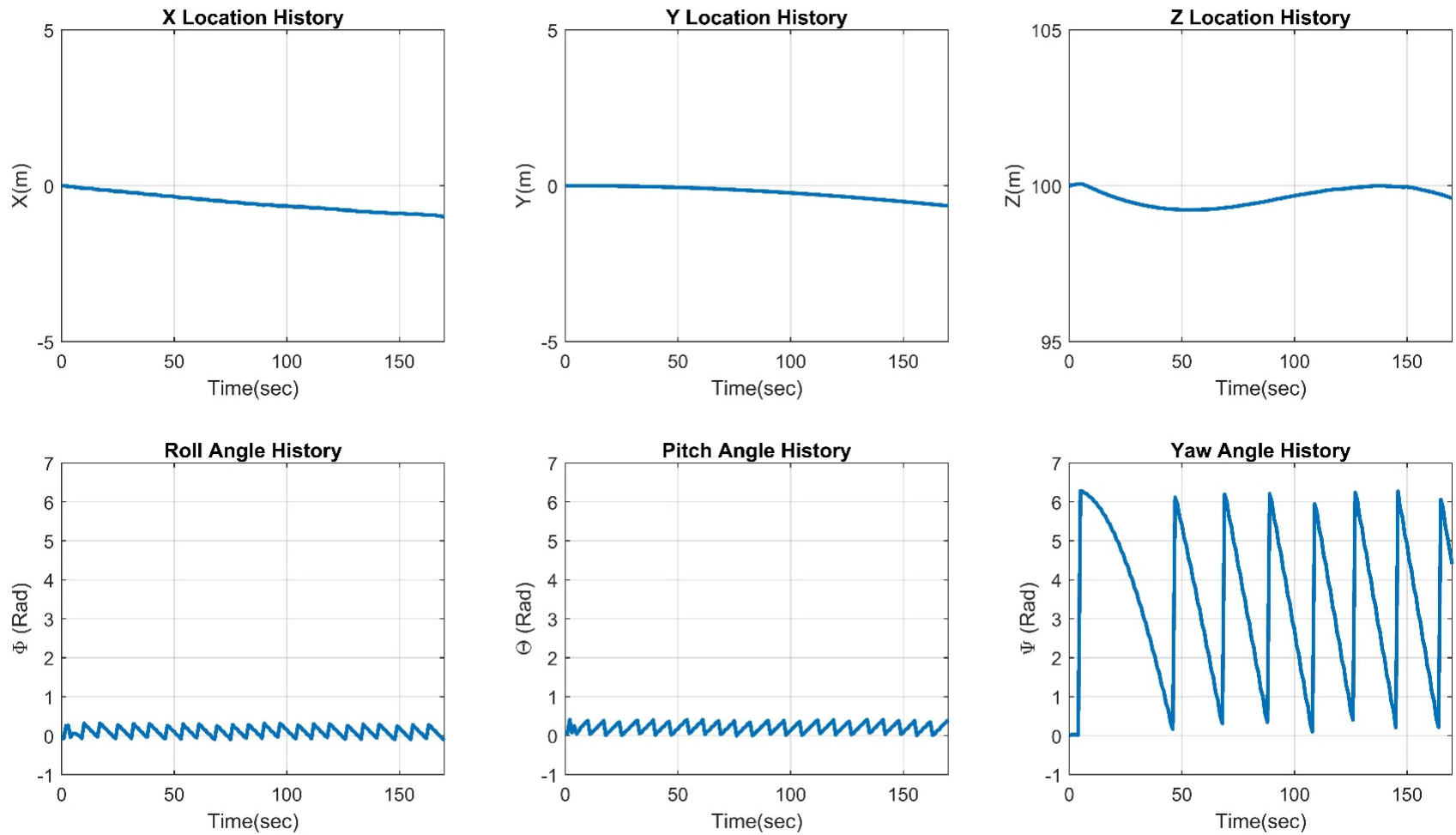

Figure 10. This figure shows the state's history for an episodic length of 180 secs using DDPG Algorithm

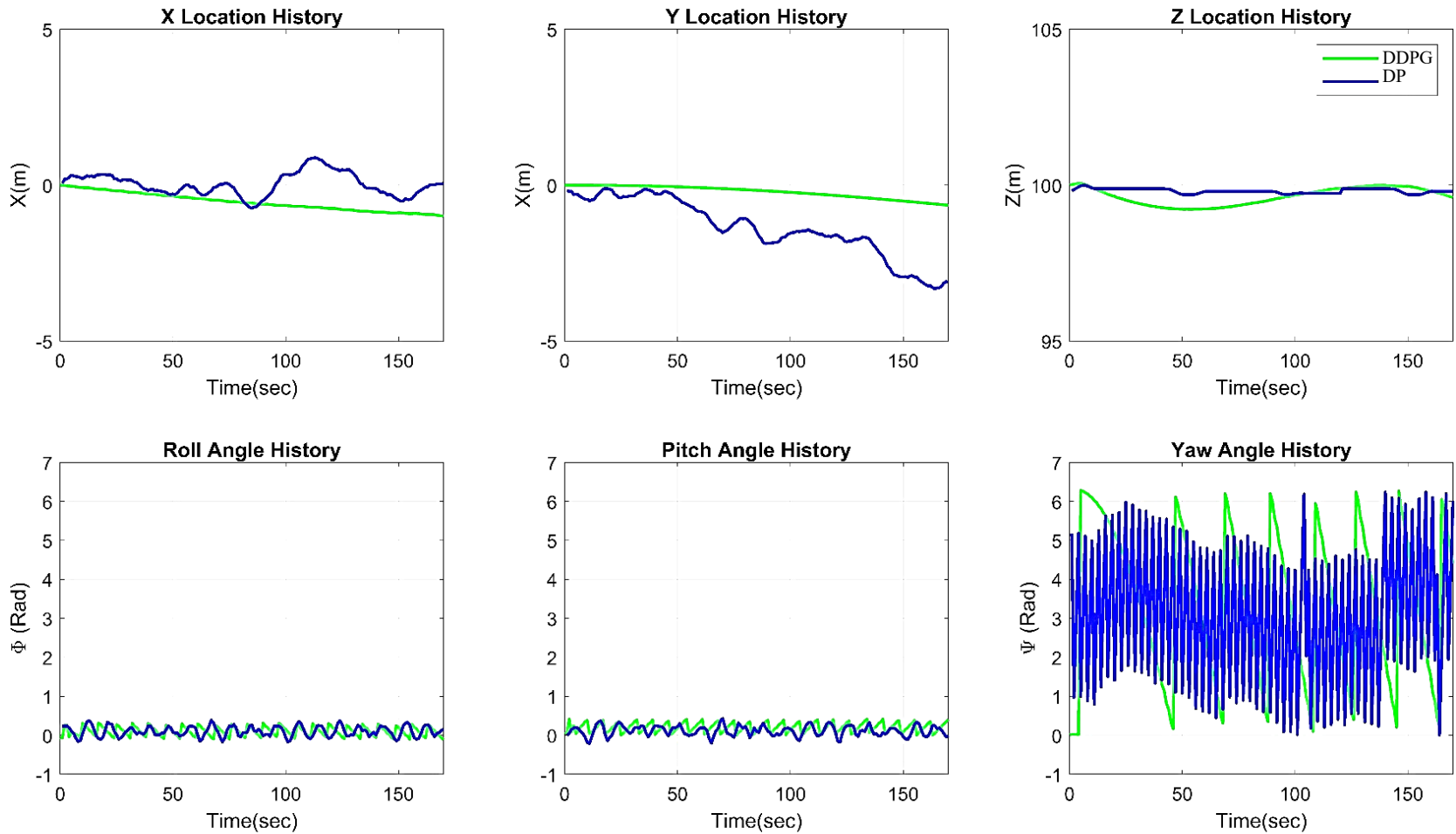

**Figure 11. This figure shows the comparison between the results of Dynamic Programming Algorithm and DDPG Algorithm.**

C.  *Different Initial Conditions Results*:   To verify the robustness of the RL controller designed by model free dynamic programming code, we test it with different initial conditions. The propeller might fail at any translational velocity or any orientation. Keeping in view nature of the problems the following set of initial conditions were given to the DP RL controller.

Table 3. Different Initial Conditions

| State | Initial disturbance in State | Method | Resultant Location Coordinates | | | | | |
|---|---|---|---|---|---|---|---|---|
| | | | Xmin (m) | Xmax (m) | Ymin (m) | Ymax (m) | Zmin (m) | Zmax (m) |
| IC 1 | U = 5 m/s | DP | 0 | 12 | -8 | 0 | 97 | 102 |
| IC 2 | V = 5 m/s | DDPG | 0 | 7 | 0 | 15 | 98 | 102 |
| IC 3 | W = 5 m/s | DP | -12 | 2 | -14 | 3 | 90 | 100 |
| IC 4 | Phi = 0.35 rad | DDPG | -8 | 0 | 0 | 5 | 97 | 101 |
| IC 5 | Theta = 0.35 rad | DP | -15 | 0 | -4 | | 98 | 100 |

The controller was able to successfully stabilize the quadcopter for all conditions mentioned in Table 3. The result of initial condition 3 are shown in Figure 13. The quadcopter was given a downward velocity of 5 m/s in initial conditions.

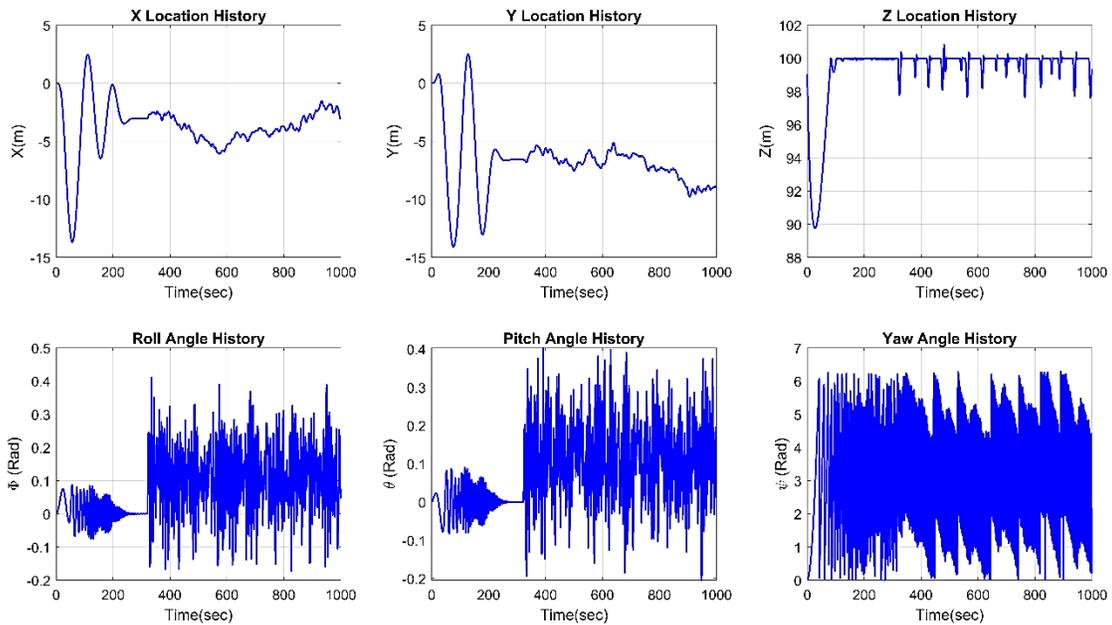

**Figure 12.  This figure shows the state history when quadcopter was given 5 m/s downward velocity in initial condition 3.**

## V. CONCLUSION

This research successfully developed a robust, nonlinear Reinforcement Learning (RL)-based control architecture for stabilizing a quadcopter following in-flight propeller failure. Despite the considerable advancements in applying RL to nonlinear systems, its application in aerospace engineering has been limited by the extensive and continuous nature of state and action spaces in aerial platforms. The RL controller developed in this study demonstrated effective stabilization of the quadcopter at a specified altitude under challenging post-failure conditions, employing two distinct algorithms: Dynamic Programming (DP) and Deep Deterministic Policy Gradient (DDPG).

Both algorithms proved effective in achieving stable hover following propeller failure. The DP algorithm generated an extended episode, sustaining hover for up to 10 minutes, while DDPG achieved a hover duration of approximately 170 seconds after training over a broad state and action space. A carefully structured reward function played a critical role in achieving this stability, ensuring the controller's ability to maintain altitude control in a post-failure configuration. Additionally, both the DP and DDPG algorithms demonstrated computational efficiency suitable for real-time applications, and simulations validated the controller's robustness across various initial conditions.

The six degrees of freedom (6 DoF) simulation further confirmed the performance of both control strategies, with the DP approach effectively enabling indefinite hover time and the DDPG agent demonstrating reliable performance in managing large, continuous state-action spaces. The neural network architecture implemented within the DDPG algorithm shows promise for extending RL-based control to other aerial platforms, particularly in tasks that require adaptive control within unknown environments. Moreover, the DDPG's scalability suggests that the RL architecture developed here may be adaptable for other dynamic systems with continuous states and multiple control inputs.

## VI. FUTURE WORK

While this study establishes a strong foundation for RL-based fault-tolerant control in quadcopters, future work will focus on extending this architecture to manage more complex failure scenarios, such as the loss of two rotors. The development of a control strategy capable of stabilizing a quadcopter under dual propeller failure would significantly enhance the scope of RL in resilient UAV control systems. Additionally, experimental testing of the proposed RL control methodologies would be performed on quadcopter. The comparison of these methods would also be compared with existing linear or nonlinear control techniques such as PID, sliding mode control or MPC etc. Experimental studies will also offer insights into the effects of environmental disturbances and model discrepancies, enabling improvements in controller adaptation and robustness. This continued research could ultimately position RL as a transformative solution in the development of fault-tolerant flight control systems across a range of UAV platforms.

# APPENDICES

*Equations of motion*

The following 6 DOF Equations of motions were used from the book "Aircraft Control and Simulation" [6], to calculate the states

(a) **Force Equations**

$$\dot{U} = RV - QW - gSin(\theta) + (Xa + Xt)/m$$
$$\dot{V} = -RU + PW + gSin(\phi)\cos(\theta) + (Ya + Yt)/m$$
$$\dot{Z} = QU - PV + gCos(\phi)Cos(\theta) + (Za + Zt)/m$$

Where U, V and W are linear velocities along X, Y and Z axis in body frame. P, Q and R are the angular rates along these axes in body frames. Xa, Ya and Za are aerodynamic forces and Xt, Yt and Zt are thrust forces. $\phi, \theta$ and $\psi$ are euler angles between body and inertial frames. As aerodynamic forces are ignored Xa, Ya and Za are set to zero. The thrust is available only in z direction so Xt and Yt are also zero. The simplified equations are as follows

$$\dot{U} = RV - QW - gSin(\theta)$$
$$\dot{V} = -RU + PW + gSin(\phi)\cos(\theta)$$
$$\dot{Z} = QU - PV + gCos(\phi)Cos(\theta) + (Fz)/m$$

Where Fz is total thrust force in z direction i.e.
$$Fz = -(F1 + F2 + F3 + F4)$$
Negative sign because Thrust forces are in upward direction and positive z-axis is pointing downward.

(b) **Kinematic Equations**

The following kinematic equations were used to calculate the Euler rates

$$Phi\_dot = P + \tan(\theta)(Qsin(\phi) + Rcos(\phi))$$
$$Theta\_dot = Q\cos(\phi) - Rsin(\phi)$$
$$Psi\_dot = (Qsin(\phi) + Rcos(\phi))/\cos(\theta)$$

(c) **Navigation Equations**

Navigation equations are used to calculate the position coordinates in the inertial axis system.

$$\dot{X}\ Loc = Uc\psi c\theta + V(-c\phi s\psi + c\psi s\phi s\theta) + W(c\psi s\phi s\theta - c\phi s\psi)$$
$$\dot{Y}\ Loc = Uc\theta s\psi + V(s\phi s\theta s\psi + c\phi c\psi) + W(c\phi s\psi s\theta - c\psi s\phi)$$
$$\dot{Z}\ Loc = -Us\theta + V(c\theta s\phi) + W(c\phi c\theta)$$

(d) **Moment Equations**

To model the moment equation, we first needed to calculate the applied moments on the quadcopter body.

$$\dot{P} = -(J_z M_x + J_{xz}M_z - J_{xz}^2 QR - J_z^2 QR + J_x J_{xz}PQ - J_{xz}J_y PQ + J_{xz}J_z PQ + J_y J_z QR)/J_{xz}^2 - J_x J_z$$
$$\dot{Q} = (M_y - J_{xz}P^2 + J_{xz}R^2 - J_x PR + J_z PR)/J_y$$
$$\dot{R} = -(J_{xz}M_x + J_x M_z + J_x^2 PQ + J_{xz}^2 PQ - J_x J_y PQ - J_x J_{xz}QR + J_{xz}J_y QR - J_{xz}J_z QR)/J_{xz}^2 - J_x J_z$$